\documentclass[conference]{IEEEtran}
\usepackage{amsmath,graphicx}
\graphicspath{{images/}}
\usepackage{amsfonts}
\usepackage{bm}

\ifCLASSINFOpdf
\else
\fi
\begin{document}
%
\title{Matrix Completion with Variational Graph Autoencoders: Application in Hyperlocal Air Quality Inference}

\author{
\IEEEauthorblockN{
		Tien Huu Do$^{\star \ast}$,
        Duc Minh Nguyen$^{\star \ast}$,
        Evaggelia~Tsiligianni$^{\star \ast}$,
        Angel Lopez Aguirre$^{\dagger \ddagger}$,\\
        Valerio Panzica La Manna$^{\diamond}$,
        Frank Pasveer$^{\diamond}$,
        Wilfried Philips$^{\ast \dagger}$,
        Nikos Deligiannis$^{\star \ast}$
}
\IEEEauthorblockA{\\
		$^{\star}$ Vrije Universiteit Brussel, Pleinlaan 2, B-1050 Brussels, Belgium \\
		$^{\dagger}$ IPI Ghent University, Sint-Pietersnieuwstraat 25, B-9000 Ghent, Belgium \\
		$^{\ddagger}$ Escuela Superior Polit\'{e}cnica del Litoral, Km 30.5 V\'{i}a Perimetral, Guayaquil, Ecuador \\
		$^{\ast}$ imec, Kapeldreef 75, B-3001 Leuven, Belgium \\
		$^{\diamond}$ Holst Center, imec, High Tech Campus 31, 5656 AE Eindhoven, The Netherlands
	}
}


%


\maketitle

\begin{abstract}
Inferring  air quality  from a limited number of observations is an essential task for monitoring and controlling air pollution. Existing inference methods typically use low spatial resolution data collected by fixed monitoring stations and infer the concentration of air pollutants using additional types of data, e.g., meteorological  and traffic information. In this work, we focus on street-level air quality inference by utilizing data collected by mobile stations. We formulate air quality inference in this setting as a graph-based matrix completion problem 
and propose a novel variational model based on graph convolutional autoencoders. Our model  captures effectively the spatio-temporal correlation of the measurements and does not depend on the availability of  additional information apart from the street-network topology. Experiments on a real air quality dataset, collected with mobile stations, shows that the proposed model outperforms state-of-the-art approaches.
\end{abstract}
%
%
%
%
\section{Introduction}
\label{sec:intro}

Air pollution is one of the most serious threats for the human health and the environment. 
In order to mitigate air pollution, we need to accurately  
measure air quality at very high spatial and temporal rates, 
especially within urban areas. 
Fixed monitoring stations have been deployed to measure the concentration of air pollutants. Given the high cost of the necessary  instruments, the number of such installations is  limited. 
Although  fixed stations can collect measurements with high temporal resolution, their  spatial resolution  is very low; hence, there is a need to spatially infer the concentration of air pollutants.  
Recent advances in sensors, IoT platforms, and mobile communications enable deploying low-cost mobile monitoring stations, e.g., by mounting sensors on vehicles. 
Examples include the air quality monitoring system using the public transport network in Zurich~\cite{hasenfratz2014pushing}, the system using Google street-view cars in Oakland, CA~\cite{apte2017high}, and imec's City-of-Things platform that uses postal trucks~\cite{latre2016city}. 
Deploying mobile  stations increases the spatial density of air quality  measurements; 
however, their temporal resolution per location is low since the vehicles are moving. In addition, there are still locations not covered by the vehicles. 
This renders computationally inferring missing air quality measurements across the spatial and temporal dimensions a problem of high interest. 

A number of methods have been proposed to infer the air pollutant concentration using measurements collected by \textit{fixed monitoring stations}. 
They are based on either \textit{physical} models or \textit{data-driven} solutions~\cite{cheng2018neural}. 
In the former approach, the complex physical dispersion processes of air pollutants are modeled using observed data and empirical assumptions~\cite{kim2012urban,arystanbekova2004application,mensink1970aurora}. 
Methods in this category, however, often require the availability of additional information, e.g., the distribution of pollution sources and accurate weather models~\cite{airquality15}. 
Furthermore, the assumptions behind them might not hold given the variability of urban landscapes~\cite{cheng2018neural}. 
Data-driven methods do not rely on strong assumptions; 
instead, they utilize diverse local data, such as meteorological information, points of interest and traffic information, to infer the concentration of air pollutants. 
By leveraging the recent advances in deep learning, in particular, data-driven methods have achieved good inference performance~\cite{cheng2018neural,qi2018deep,fan2017spatiotemporal}.

Only very limited work has focused on  air quality inference using data collected by mobile  stations~\cite{hasenfratz2014pushing}. 
In this paper, we use the City-of-Things platform from imec \cite{latre2016city} to retrieve street-level air quality data measured using mobile stations in Antwerp, Belgium. 
Given the available data, we infer the air quality in unmeasured locations across time and space. 
We follow a data-driven approach and formulate the air quality inference problem as a graph-based matrix completion problem. Specifically, we exploit the topology of Antwerp'’s street network and propose a novel deep learning model based on variational graph autoencoders; we refer to our model as AVGAE. 
The model captures effectively the spatio-temporal dependencies in the measurements, without using other types of data, such as traffic or weather, apart from the street-network topology. 
Experiments on  real data from the City-of-Things platform 
show that our method outperforms various reference models.

To summarize, our main contributions in this paper are: (\textit{i}) we formulate air quality inference as a graph-based matrix completion problem 
and propose a variational graph autoencoder for accurate inference. To the best of our knowledge, this is the first work to explore graph-based neural network models in the context of air quality inference; 
(\textit{ii}) the proposed model effectively incorporates the temporal and spatial correlations 
via a temporal smoothness constraint and graph convolutional operations; (\textit{iii}) we carry out comprehensive experiments on real-world datasets to evaluate the proposed model showing  its superior performance compared to  existing models. 

The rest of this paper is as follows: Section~\ref{sec:related_work} reviews the related work, and Section~\ref{sec:method} states the problem and presents our model. Section~\ref{sec:experiments} describes the experiments and Section~\ref{sec:conclusion} concludes the paper.

\section{Related Work}
\label{sec:related_work}


\subsection{Air Quality Inference}

Unmeasured air pollution in locations or time instances can be estimated using simple interpolation or resampling techniques~\cite{fan2017spatiotemporal,li2011spatiotemporal}. However, given the dynamics of air pollutants, these techniques tend to produce high estimation errors. 
Alternatively, one can use kriging-based variogram models to capture the variance in air pollution data with respect to the geodesic distance~\cite{kitanidis1997introduction,xie2017review}. 
As a purely spatial interpolation method, however, this approach does not capture the temporal correlation in the air quality data. 

In recent years, we have witnessed the rise of machine-learning-based methods. 
In~\cite{zheng2013u}, a co-training approach with temporal and spatial classifiers is proposed for classifying  discrete air quality indices (AQIs); yet, this model can not be used to infer the real-valued concentration of air pollutants.  
Deep-neural-network-based models have been proposed for air quality inference in~\cite{cheng2018neural,qi2018deep}.  
These models exploit the spatio-temporal correlations in the concentration of air pollutants either by incorporating additional information in the model---from traffic, weather, etc.---or by imposing objective constraints. Unlike these methods, our work utilizes a graph variational autoencoder to estimate the concentration of air pollutants across space and time, and provides higher estimation performance without considering additional information. 
Alternatively, the authors of~\cite{hsieh2015inferring} proposed a model to infer the air quality using a graph-based semi-supervised approach. The work considers an affinity graph of locations and deploys a label propagation mechanism to predict the air quality. This work is similar to ours in terms of formulating the air quality inference problem on graphs; however, instead of label propagation, we propose an end-to-end graph convolutional model, which is more flexible. It is worth noting that in~\cite{qi2018deep,zheng2013u,hsieh2015inferring} the considered area is divided in a uniform grid, whereas in the proposed approach we aggregate measurements non-uniformly across the street network (see Section~\ref{sec:formulation}) 
%
\subsection{Matrix Completion on Graphs}
Matrix completion is a fundamental problem in machine learning, which focuses on inferring unknown entries of matrices~\cite{davenport2016overview}.  
Applications of matrix completion include recommender systems~\cite{nguyen2018extendable}, cellular network planning~\cite{chouvardas2016method} and air quality inference~\cite{yu2017low}, to name a few. 
Recently, a number of studies have addressed the problem of matrix completion with tools from graph signal processing~\cite{monti2017geometric,van2017graph,kalofolias2014matrix,huang2018matrix,huang2018rating} with applications in recommender systems. 
Our method is related to these approaches but it includes specific components tailored to the problem of air quality inference from mobile measurements. In the experimental section, we compare the performance of our method against~\cite{nguyen2018extendable,monti2017geometric} and demonstrate its superior performance in inferring air quality data.
%
\subsection{Variational Graph Autoencoders}

Variational autoencoders (VAEs)~\cite{kingma2013auto} are generative models that have lately received considerable attention. The study in~\cite{liang2018variational} proposed a VAE with fully connected neural network layers with application in collaborative filtering, a particular application of matrix completion. Furthermore, variational inference on graphs has been proposed for link prediction~\cite{kipf2016variational}. Our model is different from~\cite{kipf2016variational,liang2018variational} in that we propose a variational graph autoencoder, which can express the spatial and temporal dependencies across air pollution measurements. Furthermore, the data in~\cite{liang2018variational} is assumed to follow a discrete multinomial distribution, whereas in our model, the data follows a continuous distribution.

\section{Method}
\label{sec:method}

\begin{figure*}[t]
\centering
\includegraphics[width=0.9\textwidth]{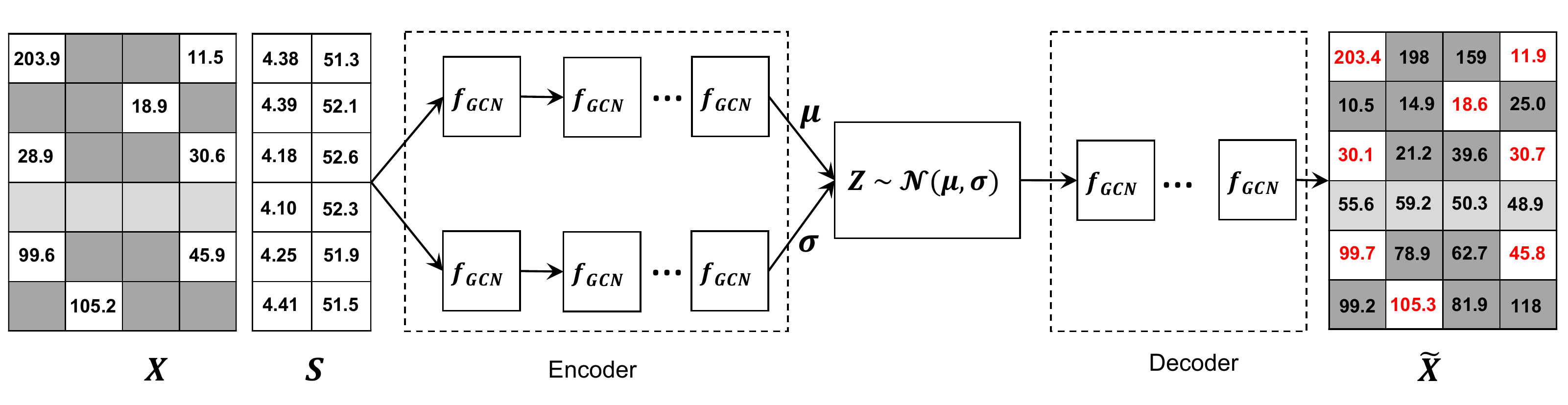}
\caption{The proposed variational graph autoencoder architecture for air quality inference (AVGAE). 
The input of AVGAE consists of the incomplete matrix $\bm{X}$ and the matrix of geocoordinates $\bm{S}$. 
The light gray row in $\bm{X}$ indicates a location without  measurements across time, dark gray cells represent unmeasured locations at a given time instance, and the entries with a red font are reconstructed known entries on which we evaluate the loss function. 
The function blocks $\bm{f}_{\text{GCN}}$ represent GCN layers. The encoder outputs the parameters $\bm{\mu}$, $\bm{\sigma}$ of a Gaussian distribution. The output matrix $\tilde{\bm{X}}$ approximates the known entries and contains the inferred unknown entries.}
\label{fig:model1}
\end{figure*}

\subsection{Problem Formulation and Notation}
\label{sec:formulation}

We focus on air quality inference at the street network of urban areas---namely, we consider only locations on streets---using measurements on the concentration of air pollutants collected by sensor-equipped vehicles moving around a specific urban area; the problem statement adheres to the smart cities concept.  Each vehicle makes measurements while moving on the city street network, resulting in high spatial measurement density; in contrast, the measurements at a specific location have low temporal resolution.

As the time and location associated to a measurement are continuous, 
it is  convenient to aggregate the measurements at discrete time instances  and locations. 
We uniformly divide the time span of the data into equal slots of duration $\tau$ (e.g., one hour). 
In a given timeslot $t$, we gather all measurements within a pre-defined geographical distance $r$ from a given spatial location $p$ on the street network and take their median-value as the measurement at  location $p$ at timeslot $t$. 
The street network information is obtained from OpenMapTiles\footnote{https://openmaptiles.com/downloads/europe/belgium/antwerp/}. Hence, the aggregation across space is non-uniform and is adapted to the considered locations on the street network. 

The above aggregation process results in a measurement matrix $\bm{X} \in \mathbb{R}^{N \times T}$, 
with $N$ the number of considered geographical locations and $T$ the number of timeslots. 
An entry $\bm{X}_{ij}$, with $i = 1,\dots,N$ and $j = 1,\dots,T$, corresponds to the measurements 
at the $i^{\text{th}}$ location and the $j^{\text{th}}$ timeslot. 
$\bm{X}$ is a highly incomplete matrix  with the set of known entries denoted by $\Omega$. Our task is to predict the air pollution concentration values in the unknown entries using the measurements (known entries) and the street-network topology.

For notational consistency, in the rest of the paper, we use bold-faced uppercase letters for matrices, 
bold-faced lowercase letters for vectors and regular lowercase letters for scalar variables. 
Both regular uppercase and lowercase Greek letters denote constants.

\subsection{Variational Autoencoders}


VAEs build on the assumption that the data points in a dataset can be drawn from a distribution conditioned by latent variables; furthermore, the latent variables follow a prior distribution, e.g., the Gaussian distribution. VAEs attempt to learn a deterministic function that transforms the Gaussian distribution to the distribution of the observed data.

Let $\bm{x}$ denote an example in the dataset and $\bm{z}$ the vector containing the latent variables. The inference process is modelled by

\begin{equation}
q(\bm{z} | \bm{x}) = \mathcal{N}\big( \bm{\mu},\bm{\sigma} \big),
\end{equation}

\noindent where $\bm{\mu} = f_{\mu}(\bm{x},\bm{\Theta_1})$ and $\bm{\sigma} = f_{\sigma}(\bm{x},\bm{\Theta_2})$ are parameters of the Gaussian distribution. The generative process is characterized by 
\begin{equation}
p(\bm{x} | \bm{z}) \propto f_z( \bm{z}, \bm{\Phi}).
\end{equation}

It should be noted that $f_{\mu}, f_{\sigma}$ and $f_z$ are parameterized functions and their parameters $\bm{\Theta_1}, \bm{\Theta_2}$ and $\bm{\Phi}$ can be learned from data. These functions are often implemented by neural network layers. 
To find the parameters, one needs to minimize the following equation:

\begin{equation}\label{core}
\mathcal{L} = -\mathbb{E}_{q(\bm{z} \vert \bm{x})} \big[ \text{log}p(\bm{x}|\bm{z}) \big] + \mathcal{D}\big[ q(\bm{z}|\bm{x}) \lVert p(\bm{z}) \big].
\end{equation}

In~\eqref{core}, one can interpret the first term as the reconstruction error and the second term as a regularization constraint. The second term is the Kullback-Leibler (KL) divergence between $q(\bm{z}|\bm{x})$ and the prior $p(\bm{z}) = \mathcal{N}(\bm{0},\bm{I})$, which can be computed with a closed form formula~\cite{kingma2013auto}.

\subsection{Variational Graph Autoencoders}\label{gvae}
Variational graph autoencoders (VGAEs)~\cite{kipf2016variational} adhere to the VAE concept and utilize graph convolutional layers (GCN) for the parameterized functions $f_{\mu}, f_{\sigma}$ and $f_z$. 
Given a graph $\mathcal{G} = (\mathcal{V},\mathcal{E})$ with an adjacency matrix $\bm{A} \in \mathbb{R}^{N \times N}$ and a degree matrix $\bm{D} \in \mathbb{R}^{N \times N}$, $N = \vert \mathcal{V} \vert$, a graph convolutional layer~\cite{kipf2016semi} is expressed as

\begin{equation}
f_{\text{GCN}}(\bm{X}) = \sigma \big( \tilde{\bm{D}}^{-\frac{1}{2}} \tilde{\bm{A}} \tilde{\bm{D}}^{-\frac{1}{2}} \bm{X} \bm{W} \big)
\end{equation}

\noindent where $\tilde{\bm{A}} = \bm{A} + \bm{I}_N$, $\tilde{\bm{D}}_{ij} = \sum_j \tilde{\bm{A}}_{ij}$, $\bm{X} \in \mathbb{R}^{N \times T}$ is the input signal summarized in a matrix, $\bm{W} \in \mathbb{R}^{T \times D}$ is the corresponding weight matrix with $D$ being the GCN layer's dimensionality, and $\sigma$ indicates a nonlinear function. By stacking multiple GCN layers, more complex functions can be constructed. In what follows, we  propose a particular architecture tailored to the air quality inference task.

\subsection{The Proposed AVGAE Architecture}
The architecture of our model, which we refer to as AVGAE, is depicted in Fig.~\ref{fig:model1}.
We build a graph of $N$ nodes by considering the geodesic distance among the $N$ corresponding discretized locations on the  street network. Two nodes are connected if the geodesic distance between them is smaller than a predefined threshold $\delta$, or if they belong to the same road segment. The weight of a  connection is the inverse of the geodesic distance in meters computed by the Haversine formula~\cite{van2012heavenly}. Furthermore, we summarize the locations' geocoordinates in a matrix $\bm{S}$. Our model is described by the following set of equations:
\begin{align}
\bm{\mu} &= \text{GCN}_{\mu}(\bm{X, S, \Theta_1} ) \label{eq:gcn_mu} \\
\bm{\sigma} &= \text{GCN}_{\sigma}(\bm{X, S, \Theta_2} ) \label{eq:gcn_sigma} \\
\bm{Z} &\sim \mathcal{N} (\bm{\mu},\bm{\sigma}) \label{eq:z} \\
\bm{\tilde{X}} &= \text{GCN}_z(\bm{Z}, \bm{\Phi}) \label{eq:gcn_x}
\end{align}

In~\ref{eq:gcn_mu},~\ref{eq:gcn_sigma},~\ref{eq:gcn_x}, $\text{GCN}_{\mu}$, $\text{GCN}_{\sigma}$ and $\text{GCN}_z$ are functions obtained by stacking GCN layers and, $\bm{\Theta}_1$, $\bm{\Theta}_2$ and $\bm{\Phi}$ are parameters that can be learned from the data. 
$\bm{S}$ is the geocoordinates matrix, which is horizontally concatenated with $\bm{X}$.  
Our model utilizes two separate branches for training $\bm{\mu}$ and $\bm{\sigma}$, thereby allowing to select proper activation functions for $\bm{\mu}$ and $\bm{\sigma}$ (the selected functions are mentioned in Section~\ref{sec:experiments}).

It is worth mentioning that our model is capable of inferring values at locations that are not measured by vehicles, which are illustrated by an empty row in matrix $\bm{X}$ in Fig.~\ref{fig:model1}. This is because the proposed model  captures the spatial correlation between the unobserved and observed locations through their geocoordinates and the street network's topology. 

The loss function of our model is defined in~\eqref{eq:loss}. 
We modify~\eqref{core} by using the mean absolute error (MAE) regularized by a KL divergence term. 
Even though the MAE is not everywhere differentiable, we find that using its sub-gradient is sufficient for optimization with gradient descent. 
The temporal dependency between measurements imposes an additional smoothness constraint:

\begin{multline}\label{eq:loss}
\mathcal{L}(\bm{X},\bm{\Theta}_1,\bm{\Theta}_2,\bm{\Phi}) = \frac{1}{\vert \Omega \vert} \sum_{(i,j) \in \Omega} \vert \tilde{\bm{X}}_{ij} - \bm{X}_{ij} \vert + \\ \beta \mathcal{D}\big[ q(\bm{z}|\bm{x}) \lVert p(\bm{z}) \big]  + \gamma \sum_{(i,j)} \ \sum_{k \in \mathcal{T}(i,j)} e^{-\vert j - k \vert} (\tilde{\bm{X}}_{ij} - \tilde{\bm{X}}_{i,k})^2
\end{multline}

In~\eqref{eq:loss}, $\beta$ and $\gamma$ are positive tuning parameters and $\mathcal{T}(i,j)$ is the neighborhood of the entry $\bm{X}_{i,j}$ with respect to the temporal dimension. 
The width of the neighborhood $w_{\mathcal{T}}$ is a parameter that is fine-tuned experimentally. 
We minimize the loss function with respect to training entries using the stochastic gradient descent---where we use the reparameterization technique in~\cite{kingma2013auto}---and we deploy the dropout regularization technique to mitigate overfitting. 
After training, we obtain the re-constructed data matrix $\tilde{\bm{X}}$ containing predicted values for the unknown entries. 

\section{Experiments}
\label{sec:experiments}

\begin{table*}[t]
\centering
\caption{Air quality inference results.}
\label{table:classification_result}
 \begin{tabular}{c | c c | c c} 
 \hline \hline
  & \multicolumn{2}{|c|}{$\text{NO}_2$} & \multicolumn{2}{|c}{$\text{PM}_{2.5}$} \\
  \hline
  & MAE  & RMSE & MAE  & RMSE \\
\hline
\hline
Kriging linear~\cite{kitanidis1997introduction} & 18.19 & 28.43 & 3.28 & 7.98 \\
\hline
Kriging exponential~\cite{kitanidis1997introduction} & 15.86 & 25.58 & 2.89 & 7.43 \\
\hline
KNN-based collaborative filtering~\cite{koren2010factor} & 20.92 & 32.67 & 3.60 & 7.47 \\
\hline
SVD~\cite{mnih2008probabilistic} & 27.35 & 38.32 & 7.41 & 13.40 \\
\hline
NMF~\cite{luo2014efficient} & 71.67 & 82.34 & 6.75 & 13.09 \\ 
\hline
NMC~\cite{nguyen2018extendable} & 22.12 & 32.83 & 3.99 & 8.35 \\
\hline
RGCNN~\cite{monti2017geometric} & 48.6 & 60.11 & 6.2 & 15.4 \\
\hline \hline
AVGAE (Our method) &  \textbf{14.92} & \textbf{24.33} & \textbf{2.56} & \textbf{6.42} \\
\hline \hline
\end{tabular}
\end{table*}

\subsection{The Dataset}

We rely on the City-of-Things platform~\cite{latre2016city} to obtain measurements of the air quality in the Antwerp city in Belgium. The platform makes use of 24 cars equipped with mobile monitoring devices. We retrieve the measurements during May 2018 for two air pollutants, that is, $\text{NO}_2$ and $\text{PM}_{2.5}$. 

As described in Section~\ref{sec:formulation}, we first apply aggregation as a data preprocessing step, where we choose $\tau = 1$ hour and $r = 100$ meters. It is worth mentioning that these parameters can be made smaller, leading to near real-time inference with a finer spatial resolution. After processing, we obtain 3630 and 4086 discrete locations for the $\text{NO}_2$ and $\text{PM}_{2.5}$ datasets, respectively. Each location is specified by the pair of latitude and longitude geocoordinates. Moreover, for each pollutant, a location is associated with a measurement vector of $\text{T} = 30 \times 24 = 720$ dimensions, which is the number of hours during the considered period. The description of the dataset is presented in Table~\ref{table:dataset}.

\begin{table}[t] 
\centering
\caption{The description of the $\text{NO}_2$ and $\text{PM}_{2.5}$ dataset. 
The units for $\text{NO}_2$ and $\text{PM}_{2.5}$ are parts per billion (ppb) and $\mu \text{G}/\text{m}^3$.
}
\label{table:dataset}
 \begin{tabular}{ c | c | c }
 \hline \hline
  & $\text{NO}_2$ & $\text{PM}_{2.5}$  \\
  \hline \hline
 Number of locations & 3630 & 4086 \\ 
 \hline
 Duration in hours  & 720 & 720  \\
 \hline
 Max concentration  & 633.65 & 189.03  \\
 \hline
 Min concentration & 0.16 & 0.07 \\
 \hline
 Mean concentration & 85.50 & 9.83 \\
 \hline
 \% of known entries versus all & 0.60 & 0.56 \\
 \hline \hline
\end{tabular}

\end{table}

%

\subsection{Experimental Setting}

To evaluate the proposed method, we randomly divide the known entries into  training and test sets. That is, 90\% of the known entries is used for training and the rest is reserved for testing. We use two common evaluation metrics, namely, the root mean squared error (RMSE) and the mean absolute error (MAE). To obtain  robust results, we repeat this procedure with 5 random divisions and report  average   results.


To create the graph, we set the distance threshold to $\delta = 200$ m. The parameters of the AVGAE are chosen experimentally: we set the learning rate to $\alpha = 0.005$, the KL divergence coefficient to $\beta = 0.1$, the temporal smoothness coefficient to $\gamma = 0.8$, the temporal neighborhood width to $w_{\mathcal{T}} = 3$ and the dropout rate to 0.4. For all GCN layers, we use the same dimensionality, that is,  $D = 512$. 
We use 4 GCN layers for the encoder and 1 GCN layer for the decoder. We employ ReLU to activate the GCN layers of the \text{encoder} except for the last GCN layer of the $\bm{\sigma}$ branch where the sigmoid function is used because $\bm{\sigma}$ should contain strictly positive entries. Because the output is unbounded, it is not necessary to use an activation function for the GCN layer of the decoder.

As reference benchmarks, we have selected two well-established kriging-based models, that is, the linear and exponential models~\cite{kitanidis1997introduction}. A kriging model is applied per column  of the matrix~$\bm{X}$ (corresponding to a timeslot) using the geocoordinates information in~$\bm{S}$.    Furthermore, we consider various state-of-the-art matrix completion methods, including KNN-based collaborative filtering~\cite{koren2010factor}, SVD-based matrix completion~\cite{mnih2008probabilistic}, non-negative matrix factorization~\cite{luo2014efficient}, and extendable neural matrix completion~\cite{nguyen2018extendable}. These models perform completion under an assumption on~$\bm{X}$, e.g., a low-rank prior. Furthermore, we compare against the graph-based matrix completion method in~\cite{monti2017geometric}; specifically, the RGCNN model, where the graph for  the row-factor matrix is the same as in  our AVGAE model and  the hyper-parameters are kept as in~\cite{monti2017geometric}.  For the implementation, we rely on PyKridge\footnote{https://pykrige.readthedocs.io/en/latest/index.html} for the kriging models and Surprise\footnote{https://surprise.readthedocs.io/en/stable/index.html} for the reference matrix completion techniques. The implementations of~\cite{nguyen2018extendable,monti2017geometric} are available online.
All models
have been trained in our dataset.

\subsection{Result and Analysis}
The results in air quality inference with the different methods are shown in Table~\ref{table:classification_result}. Kriging-based methods provide good estimation accuracy, particularly the exponential model. This is because such models capture properly the spatial correlation in the  air
quality measurements with respect to the geodesic distance.

On the other hand,  matrix completion models assume that there are
hidden factors characterizing rows (a.k.a., discrete locations) and columns (a.k.a., timeslots). 
While this assumption is appropriate for other problems such as recommendation systems, it does
not properly capture the spatio-temporal correlation in the concentration of air pollutants. 

It is evident that our  AVGAE model achieves the best performance for both the RMSE and MAE metrics and for both pollutants ($\text{NO}_2$ and $\text{PM}_{2.5}$). 
Conversely to kriging models, AVGAE effectively captures \textit{both} the temporal and spatial correlations in the data, and leverages the underlying graph structure of the street network. Furthermore, unlike the reference matrix completion models, either graph-based or not, AVGAE\ adheres to an autoencoder model, which provides good performance in reconstruction problems.

\section{Conclusion}
\label{sec:conclusion}


Measuring the concentration of air pollutants with mobile  stations is a promising approach to achieve hyperlocal air quality monitoring. The measurements collected by such mobile stations, however, have very low temporal resolution per location and there are still unmeasured locations. 
We formulated the air quality inference problem in this setting as a matrix completion problem on graphs, and proposed a variational graph autoencoder model to solve it. The proposed model was experimentally shown to effective capture the spatio-temporal correlation in the measurements, resulting in better air quality inference compared to various state-of-the-art kriging and matrix completion methods.



\bibliographystyle{IEEEbib}
\bibliography{refs}

\end{document}